\documentclass[conference]{IEEEtran}
\IEEEoverridecommandlockouts
\usepackage{cite}
\usepackage{amsmath,amssymb,amsfonts}
\usepackage{algorithmic}
\usepackage{graphicx}
\usepackage{textcomp}
\usepackage{xcolor}
\def\BibTeX{{\rm B\kern-.05em{\sc i\kern-.025em b}\kern-.08em
    T\kern-.1667em\lower.7ex\hbox{E}\kern-.125emX}}
\begin{document}

\title{Watch the Speakers: A Hybrid Continuous Attribution Network for Emotion Recognition in Conversation With Emotion Disentanglement

\thanks{*Corresponding authors. This work was supported in part by the National Natural Science Foundation of China under Grant 62236005, 61936004, and U1913602.}
}


\author{
\tiny
\IEEEauthorblockN{1\textsuperscript{st} Shanglin Lei\textsuperscript{12}, 2\textsuperscript{nd} Xiaoping Wang\textsuperscript{12*}, 3\textsuperscript{rd} Guanting Dong\textsuperscript{4}, 4\textsuperscript{th} Jiang Li\textsuperscript{123*}, 5\textsuperscript{th} Yingjian Liu\textsuperscript{12}}
\IEEEauthorblockA{
\textsuperscript{1}\textit{School of Artificial Intelligence and Automation}, \textit{Huazhong University of Science and Technology} , Wuhan 430074, China \\
\textsuperscript{2}\textit{Key Laboratory of Image Processing and Intelligent Control of Education Ministry of China}, Wuhan 430074, China\\
\textsuperscript{3}\textit{Institute of Artificial Intelligence, Huazhong University of Science and Technology (HUST)}, Wuhan 430074, China\\
\textsuperscript{4}\textit{School of Artificial Intelligence, Beijing University of Posts and Telecommunications}, Beijing 100876, China \\
\{lawson, wangxiaoping, lijfrank, M202072868\}@hust.edu.cn, dongguanting@bupt.edu.cn
}
}

\maketitle

\begin{abstract}
 Emotion Recognition in Conversation (ERC) has attracted widespread attention in the natural language processing field due to its enormous potential for practical applications.
 Existing ERC methods face challenges in achieving generalization to diverse scenarios due to insufficient modeling of context, ambiguous capture of dialogue relationships and overfitting in speaker modeling.
In this work, we present a Hybrid Continuous Attributive Network (HCAN) to address these issues in the perspective of emotional continuation and emotional attribution.
Specifically, HCAN adopts a hybrid recurrent and attention-based module to model global emotion continuity. Then a novel Emotional Attribution  Encoding (EAE) is proposed to model intra- and inter-emotional attribution for each utterance.
Moreover, aiming to enhance the robustness of the model in speaker modeling and improve its performance in different scenarios, 
A comprehensive loss function emotional cognitive loss $\mathcal{L}_{\rm EC}$ is proposed to alleviate emotional drift and overcome the overfitting of the model to speaker modeling.
Our model achieves state-of-the-art performance on three datasets, demonstrating the superiority of our work.
Another extensive comparative experiments and ablation studies on three benchmarks are conducted to provided evidence to support the efficacy of each module. 
Further exploration of generalization ability experiments shows the plug-and-play nature of the EAE module in our method.

\end{abstract}

\begin{IEEEkeywords}
Natural language processing, emotion recognition in conversation, 
context modeling, dialogue relationship
\end{IEEEkeywords}
\section{Introduction}
\begin{figure}[htbp]
\centerline{\includegraphics[scale=0.5]{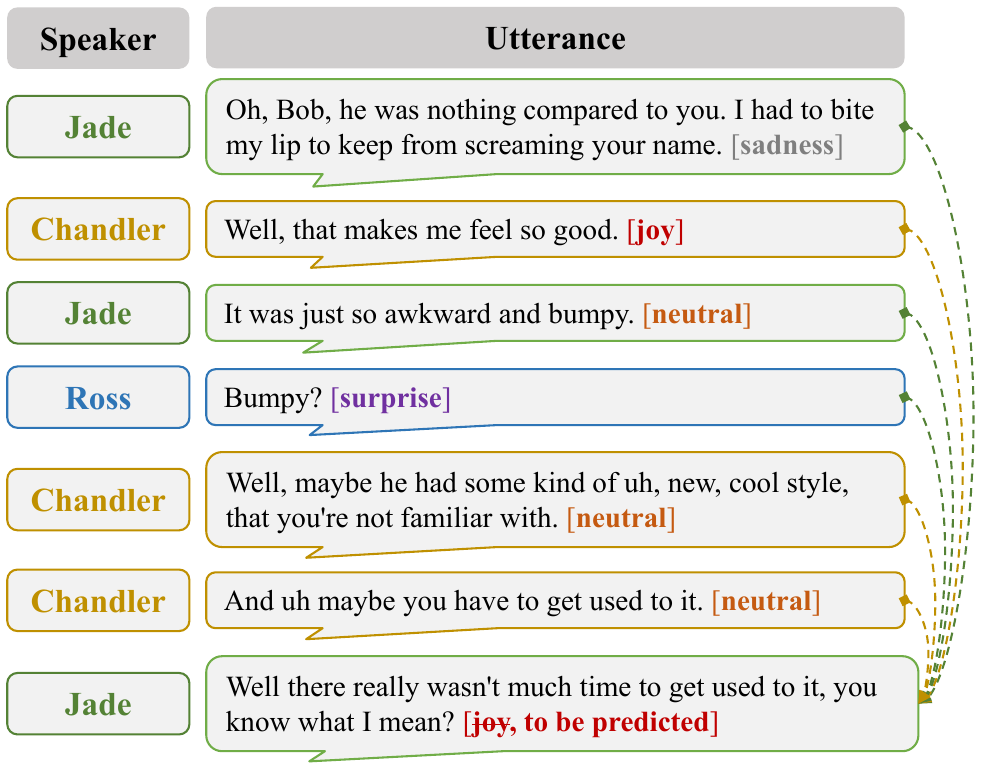}}
\caption{A example for the conversation in the MELD dataset.}
\label{fig:example}
\end{figure}

Emotion Recognition in Conversation (ERC) is a rapidly growing research field within Natural Language Processing (NLP) that focuses on identifying the emotions conveyed in each utterance of a conversation. 
Different from the single sentence's emotional classification in explicit sentiment analysis\cite{munikar2019fine,yin2020sentibert,do2019deep,sun2019utilizing}, this task contains samples with vastly different conversation lengths, ambiguous emotional expressions, and complex conversational relationships. Fig.~\ref{fig:example} illustrates an example of the conversation scenario, where the utterance to be predicted (the last utterance) is influenced by the historical utterances of that conversation.
As expected, ERC task has attracted the attention of many researchers due to its potential applications in various fields such as political campaigning and public opinion analysis\cite{cambria2017sentiment, anstead2015social}, human-robot interaction\cite{sheridan2016human} and task-oriented dialogue system \cite{rashkin2018towards, qixiang-etal-2022-exploiting,zeng2022semi,sun2023improving}.

Pervious ERC methods generally formulate the task as a supervised learning task based on different architectures of neural networks. This places a significant demand on the model's ability to capture the context of each utterance and effectively utilize speaker information \cite{shen2021directed}.
Moreover, various modeling methods for context and speaker have significantly raised the baseline, but there are still two remaining challenges of ERC need to solve. 
(1) {\bf Insufficient modeling of context.} 
Existing works on context modeling can be broadly categorized into two types:
The recurrent based methods\cite{majumder2019dialoguernn,ghosal2020cosmic,hazarika2018icon,jiao2019higru} focus on establishing more natural context temporal correlation. 
However, these methods may struggle to capture the global emotional continuity in long conversations.
Although attention-based methods \cite{zhong2019knowledge,ghosal2019dialoguegcn,shen2021dialogxl,li2022contrast} aim to aggregate emotional features at multiple levels, they may not be as effective as temporal models in capturing emotional continuity between speakers over time.
These methods adopt a single and redundant network architecture, which results in a lack of generalization in context modeling. 
(2) {\bf Ambiguous capture of dialogue relationships.} 
Studies \cite{chen2023dialogue,sun2021discourse} provide evidence that generating emotional responses can effectively improve the performance of ERC models. It can be inferred that in real-life conversations, more direct conversational relationships often lead to more direct emotional transmission.
Nonetheless, the ERC field still lacks of detailed modeling of the emotional influence within and between speakers in the perspective of dialogue relationship.
(3) \textbf{Overfitting in speaker modeling.}
In the ERC task, speakers often exhibit distinct characteristics in their emotional expressions due to differences in identity and personality. To better leverage fine-grained information, several studies have made significant contributions \cite{dong2023multi,liu2020coach}. 
Although intricate network designs have been developed from various perspectives, such as speaker psychological states, dialogue memory, and relative positional relationships, these approaches have yielded limited results.
Specifically, The models have encountered overfitting issues in different dialogue scenarios, which has hindered their effectiveness.
Therefore, these three limitations greatly hinder the application of ERC models in real-world scenarios, which is precisely what our work aims to address.


We have proposed HCAN to effectively address the aforementioned issues. 
To tackle the problem of insufficient context modeling, we propose Emotional Continuation Encoding (ECE) to extract more robust features in different conversation situations, which comprehensively utilizes both the recurrent units and the attention blocks. 
The \textit{Attribution Theory}\cite{schachter1962cognitive} proposes that a stimulus triggers perception, which leads individuals to consider the situation, and physiological reactions lead to cognitive interpretation of physiological changes, both of which together result in emotional expression.
Drawing inspiration from the \textit{Attribution Theory} and accurately capturing dialogue relationships, we present Emotional Attribution  Encoding (EAE) based on IA-attention, which models the intra-attribution and inter-attribution of each sentence in an attribution perspective. 
Due to the diverse input perturbations in conversations \cite{dong2022pssat,guo2023revisit}, we also design emotional cognitive loss to effectively enhance the model's robustness and extend the applicability of the overall model. 
The Emotional Cognitive loss $\mathcal{L}_{EC}$ is composed of cross-entropy $\mathcal{L}_{cross}$, KL divergence $\mathcal{L}_{\rm KL}$ for predicting and recognizing emotions, and Adversarial Emotion Disentanglement loss $\mathcal L_{adv}$ \cite{li2022robust}. Among them, cross-entropy calculation serves as the main emotional loss, KL divergence can alleviate emotional drift, and Adversarial Emotion Disentanglement loss can mitigate the overfitting of the model to speaker modeling.

Our contributions are three-fold:

(1) By combining the recurrent and attention-based approaches, our proposed ECE module achieves strong robustness in global emotion continuity modeling across different datasets, particularly demonstrating outstanding performance on long conversation samples.

(2) Consider capturing dialogue relationships in the perpective of \textit{Attribution Theory}, we propose an original IA-attention to extract intra-attribution and inter-attribution features, which offers a more direct and accurate modeling of human emotional comprehension.

(3) Our model achieves state-of-the-art performance on three datasets, demonstrating the superiority of our work. The proposed EAE module is a plugin module that exhibits strong generalization and effectiveness across different baselines.





\begin{figure*}[htbp]
\centerline{\includegraphics[scale=0.48]{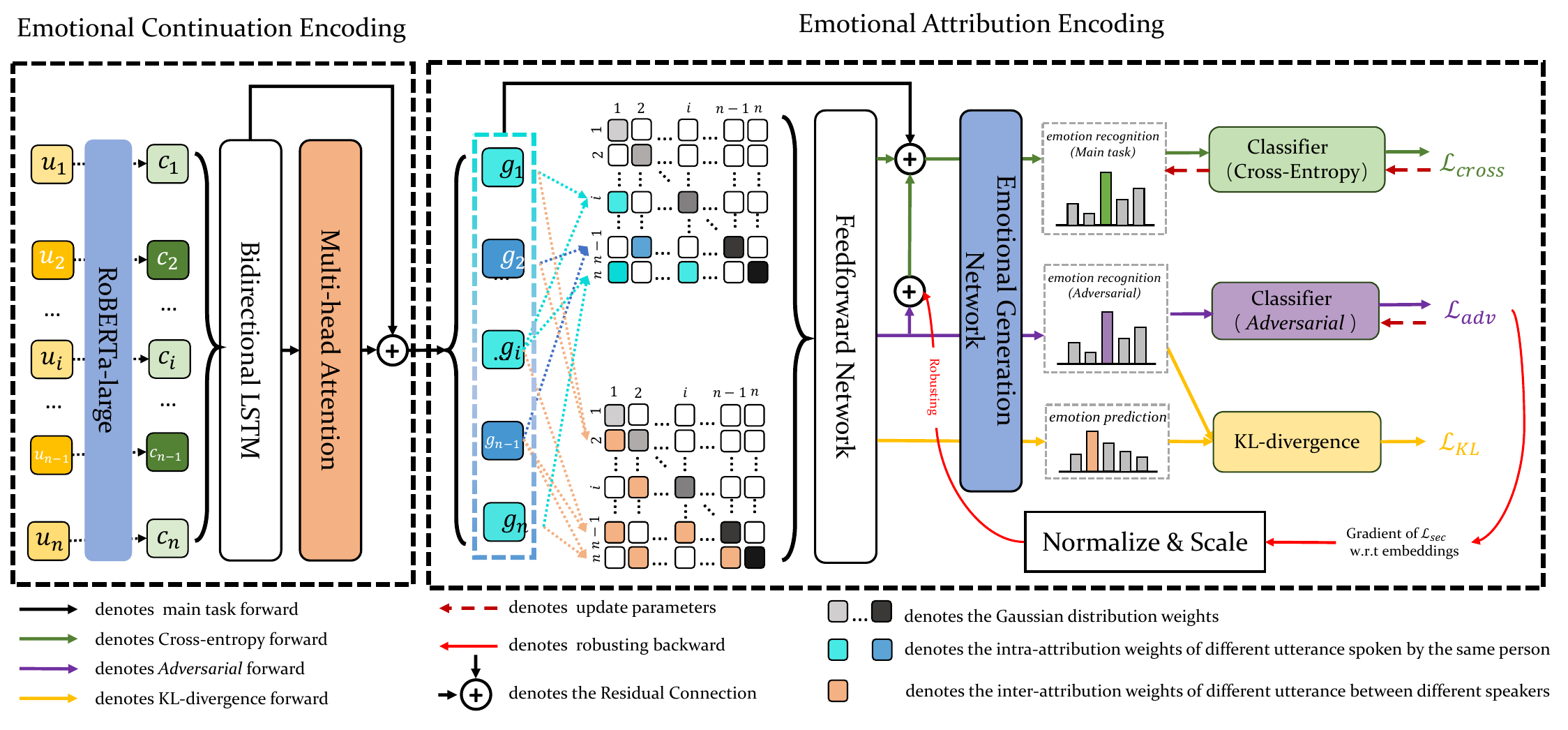}}
\caption{The overall architecture of HCAN consisting of two main components, namely Emotional Continuation Encoding and Emotional Attribution Encoding.}
\label{fig:overall}
\end{figure*}

\section{Related Work}
\subsection{Emotion Recognition of Conversation}
The significant advancement of deep learning has greatly promoted the improvement of baseline performance in ERC tasks. Recently, ERC models can be categorized into two types: recurrent-based methods and attention-based methods.
\subsubsection{Recurrent-Based Methods}
Through the use of a sequential network structure, recurrent-based methods have the potential to offer a more precise and authentic representation of the emotional dynamics present in a conversation:
DialogueRNN\cite{majumder2019dialoguernn} is the first to utilize a recurrent neural network for monitoring both speaker states and global states in conversations.
COSMIC is a conversational model that integrates commonsense knowledge to enhance its performance. This model injects commonsense knowledge into Gated Recurrent Units (GRU) to capture features related to the internal state, external state, and intent state.
The performance of SKAIG is enhanced by integrating action information inferred from the preceding context and the intention suggested by the subsequent context. 
DialogueCRN\cite{DBLP:conf/acl/HuWH20} is designed with multi-turn reasoning modules that extract and integrate emotional clues. These modules perform an iterative process of intuitive retrieval and conscious reasoning, which imitates the distinctive cognitive thinking of humans.
With the goal of achieving a comprehensive understanding of the dialogue, CauAIN \cite{zhao2022cauain} first retrieves and enhances causal clues in the dialogue through an external knowledge base. Then, it models intra- and inter-speaker interactions using GRUs.

\subsubsection{Attention-Based Methods}
To enable the extraction of emotional features at both coarse-grained and fine-grained levels, attention-based methods often employ a variety of encoders and decoders with different levels and structures.
KET \cite{zhong2019knowledge} extracts concepts related to non-pause words in neutral discourse from a knowledge base and enhances the semantic representation of vectors using a dynamic context graph attention mechanism. Finally, a hierarchical self-attention mechanism is utilized to model the dialogue level.
By leveraging four distinct attention mechanisms, DialogXL \cite{shen2021dialogxl}utilizes the language model layers of XLNet to encode multi-turn dialogues that are arranged in a sliding window.
By regarding the internal structure of dialogue as a directed acyclic graph to encode utterances, DAG-ERC offers a more intuitive approach to modeling the information flow between the distant conversation background and the nearby context.
TODKAT, as proposed in \cite{zhu2021topic}, presents a language model (LM) that is enhanced with topics through an additional layer specialized in detecting them. This model also incorporates commonsense statements obtained from a knowledge base based on the dialogue context.
\subsection{Dialogue Relation Extraction}
The task of Relationship Extraction (RE) aims to identify the relationships that exist between pairs of entities within a document. While in dialogue scenarios, the task of extracting dialogue relations becomes more challenging due to the ellipsis of expression, the fuzziness of semantic reference and the presence of long-distance context dependencies.

DialogRE \cite{yu2020dialogue} introduced the first human-annotated dataset for dialogue relationship extraction (DiaRE), which aims to capture the relationships between two arguments that arise in predictive conversations. Building upon this dataset, Chen\cite{chen2023dialogue} proposed a DiaRE method based on a graphical attention network that constructs meaningful graphs connecting speakers, entities, entity types, and corpus nodes to model the relationships between critical speakers. Similarly, Sun \cite{sun2021discourse} proposed an utterance-aware graph neural network (ERMC-DisGCN) for ERMC, which leverages a relational convolution to propagate contextual information and takes into account the self-speaker dependency of interlocutors.

Despite the promising results achieved by the aforementioned methods, they have not been validated on the ERC dataset. Furthermore, unlike directly identifying the current emotional state based on DiaRE, our approach extracts dialogue relationships from an attributional perspective and adds an emotional prediction loss to the task, which better aligns with human thought processes and enhances the robustness of the model in different scenarios.

\section{Methodology}
In this section, we present the details of how to approach conversation modeling from a continuation-attribution perspective.
The overview of HCAN is shown in Fig.~\ref{fig:overall}, which is consist of Emotional Continuation Encoding and Emotional Attribution Encoding.

\subsection{Task Statement}
In the ERC task, the goal is to identify the emotion $s_i$ of each utterance $u_i$ in a conversation $[u_1, u_2,...,u_N]$ by analyzing the dialogic context and the related speaker information $p_i$ in speaker set $\{p_i,\dots,p_M\}$, where the emotion should be selected from a pre-defined emotional target set $S$ and each utterance corresponds to one speaker in the set of speakers. 

\subsection{Emotional Continuation Encoding}
To mimic the natural conversational flow between speakers, the bidirectional LSTM is employed to encode the utterances' feature $\mathbf{c}_i \in \mathbb{R}^{d_u}$ in a temporal sequence as follows:
\[ {\bf g}^{l}_i , {\bf h}_i = \overleftrightarrow{\rm{LSTM}} ({\bf c}_i , {\bf h}_{i-1}) \tag{1} \]
where $\mathbf{h}_i \in \mathbb{R}^{2d_u}$ is the hidden state of the LSTM. Noted that the feature at the utterance-level of $u_i$ is represented by $\mathbf{c}_i \in \mathbb{R}^{d_u} $, and it is obtained through the employment of the COSMIC method for extraction. 

To avoid the vanishing of emotional continuity over long time spans, we utilized a multi-head attention module to aggregate the global information from the LSTM encoding result $\mathbf{G}^l$ as follows:
\[ \mathbf{G} = {\rm Multi\text{-}Attn} (W_Q\mathbf{G}^l, W_K\mathbf{G}^l,W_V\mathbf{G}^l) + \mathbf{G}^l  \tag{2}\]
where $\mathbf{G}^{l} = [\mathbf{g}^l_1, \dots, \mathbf{g}^l_n]$, $\mathbf{G} = [\mathbf{g}_1, \dots, \mathbf{g}_n]$, $\mathbf{g}^l_n \in \mathbb{R}^{2d_u}$, $\mathbf{g}_n\in\mathbb{R}^{2d_u}$ and $W_Q,W_K,W_V$ are trainable parameters. The use of residual connections $+$ ensures that even in the worst-case scenario, the global emotional state degrades to a temporal emotional state, thereby enhancing the robustness of the model.

\subsection{Emotional Attribution Encoding}
Emotional Attribution Encoding is the core of this article, consisting of the IA-attention module and Emotional Cognitive loss. The IA-attention module efficiently captures the dialogue relationship and establishes emotional influence from the perspective of attribution. The Emotional Cognitive loss effectively mitigates the overfitting of modeling on different datasets.
\subsubsection{IA-Attention}
Inspired by the attribution theory of emotion, we examine the emotional influnence in dialogue realtionships in an attributional prespective.
Specially, we model emotional influence as intra-attribution and inter-attribution.


To achieve this, we introduce IA-attention, which is inspired by several works about self-attention mechanism\cite{vaswani2017attention,dong2023bridging,zhao2022entity}.
This method views each global utterance representation $\mathbf{g}_i$ as a query, which is mapped to intra-attribution partial space $Q_{a}$ and inter-attribution partial space $Q_{e}$ to get two different query embeddings $\mathbf{q}_{i_a}, \mathbf{q}_{i_e}$. Meanwhile, the historical utterance $[\mathbf{g}_1,\dots \mathbf{g}_{i-1}]$ are also projected to $K$ and $V$ partial space to obtain $\mathbf{k}_i$ and $\mathbf{v}_i$.
To summarize, for each utterance, we apply different attribution attention matrices to get the intra-atribution weighted sum and inter-attribution weighted sum  which are divided by each utterance's speaker $p_i$.
The specific formula is as follows:
\[ [\mathbf{q}_{i_a};\mathbf{q}_{i_e}] = [W_{Q_a};W_{Q_e}] \mathbf{g}_i  \tag{3} \]
\[ [\mathbf{k}_1,\dots,\mathbf{k}_n] = W_{K_{\rm IA}} [\mathbf{g}_1,\dots, \mathbf{g}_n]  \tag{4} \]
\[ [\mathbf{v}_1,\dots,\mathbf{v}_n] = W_{V_{\rm IA}} [\mathbf{g}_1,\dots, \mathbf{g}_n]  \tag{5} \]
\[ \tilde{\mathbf{v}_i} = \sum_{j<i} \left( \delta_{p_j, p_i} \frac{e^{\mathbf{q}^T_{i_a} \cdot \mathbf{k}_j}}{Z} + (1-\delta_{p_j, p_i}) \frac{e^{\mathbf{q}^T_{i_e} \cdot \mathbf{k}_j}}{Z} \right) \mathbf{v}_j \tag{6}  \]

where $W_{Q_a},W_{Q_e},W_{K_{\rm IA}},W_{V_{\rm IA}} \in \mathbb{R}^{2d_u \times 4d_u}$ are trainable parameters, $\mathbf{q}_{i_a},\mathbf{q}_{i_e}, \mathbf{k}_j, \mathbf{k}_i \in \mathbb{R}^{4d_u}$ and $Z$ is the normalized factor.

To enable a more realistic perception in the dialogic relationship, the Gaussian Self-attention Mechanism\cite{guo2019gaussian} is introduced to distinguish the varying effects of dialogic temporal location. Assuming that the emotional attribution of historical utterances to the current utterance follows a normal distribution, the encoding results of the IA-attention module will be assigned weights that obey a Gaussian distribution, which is calculated as follows:
\[ \hat{\mathbf{v}}_i = \sum_{j<i}\phi(d_{i,j}|\mu ,\sigma )\tilde{\mathbf{v}}_j \tag{7}  \]
where $\hat{\mathbf{v}}_i \in \mathbb{R}^{4d_u}$,$\phi $ is a Gaussian distribution, $\mu $ and $\sigma $ are their corresponding learnable parameters, $d_{i,j}$ \cite{guo2019gaussian} stands for distance measuring the turn-taking interval between speakers.




\subsubsection{Emotional Cognitive Loss}
The emotional overfitting of the ERC task mainly focuses on emotional drift and speaker modeling. Motivated by multi-task learning \cite{li2023generative,dong2023prototypical}, our proposed Emotional Cognitive loss $\mathcal{L}_{\rm EC}$ is mainly composed of basic cross-entropy $\mathcal{L}_{cross}$, KL divergence $\mathcal{L}_{\rm KL}$ for predicting and recognizing emotions, and Adversarial Emotion Disentanglement loss $\mathcal L_{adv}$. Among them, cross-entropy calculation is the main emotional loss, KL divergence can alleviate emotional drift, and Adversarial Emotion Disentanglement loss can overcome the overfitting of the model to speaker modeling.

\textbf{Cross-Entropy Loss}  $\mathcal{L}_{cross}$, the key elements of which are computed as follows:
\[ \mathcal{D}^{\rm src}_i = {\rm Softmax}(W_{\rm D}(\lambda_\theta(\hat{\mathbf{v}}_i)+\mathbf{g}_i))  \tag{8} \]
\[ \hat{y}_{i} = {\rm Softmax}(W_o \mathcal{D}^{\rm src}_i + b_o) \tag{9} \]
\[ \mathcal{L}_{\rm cross} = - \frac{1}{\sum_{l=1}^{L}\tau (l)}\sum_{i=1}^{L}\sum_{k=1}^{\tau(i)}y_{i,k}\log(\hat{y}_{i,k})   \tag{10}\]
where $L$ is the total number of conversations in the trainset, $\tau(i)$ is the number of utterances in the conversation, $y_{i,k}$ denotes the one-hot vector and $\hat{y}_{i,k}$ denotes probability vector for candidate emotional class $n$ of the $i^{th}$ utterance in $l^{th}$ sample.

\textbf{KL Divergence} $\mathcal{L}_{\rm KL}$  are calculated as follows:
\[ \mathcal{D}^{\rm tmp}_i = {\rm Softmax}(W_{\rm D} \lambda_\theta(\hat{\mathbf{v}}_i))  \tag{11}\]
\[ \mathcal{L}_{\rm KL} = {\rm KL\textbf{-}Divergence}(\mathcal{D}^{tmp}_i, \mathcal{D}^{src}_i) \tag{12} \]

where $\lambda_\theta \in \mathbb{R}^{4d_u \times 2d_u}$ and $W_{\rm D} \in \mathbb{R}^{2d_u \times |\mathcal{E}|}$ denotes the emotional state generation network. $|\mathcal{E}|$ is the number of emotion labels.
By utilizing a shared weight matrix $W_{\rm D}$ to map the predicted emotion $\mathcal{D}^{\rm tmp}$ and the recognized emotion $\mathcal{D}^{\rm src}$, the model is able to generate more accurate emotional representations in the current emotional state and make more precise inferences based on historical utterances.


\textbf{Adversarial Emotion Disentanglement}: loss $\mathcal L_{adv}$ is proposed to further prevent the model from excessively focusing on the emotional information of a dialogue role, inspired by adversarial training methods \cite{goodfellow2014explaining,kurakin2016adversarial,manzini2019black,li2022robust,dong2022pssat}. To be more specific, given an input sentence, we obtain its hidden representations using LSTM. Next, the model classify them based on predicted probability distributions. Then, we obtain the classification cross-entropy loss $L_{cross}$. However, existing methods often being influenced by a specific dialogue role, it is difficult to consider the overall semantic information of the whole conversation, Therefore, we apply the Fast Gradient Value (FGV) technique \cite{goodfellow2014explaining,kurakin2016adversarial} to approximate the worst-case perturbation as a noise vector: 
\[ {v}_{noise} = \epsilon \frac{g}{\|g\|} ; \text{ where } g = \nabla_e \mathcal L_{cross} \tag{13} \]

Here, the gradient represents the first-order derivative of the loss function $\mathcal L_{cross}$, and $e$ denotes the direction of rapid increase in the loss function. We perform normalization and use a small $\epsilon$  to ensure the approximation is reasonable. Then, we add the noise vector $v_{noise}$ and conduct a second forward pass, obtaining a new adversarial loss $\mathcal L'_{cross}$. 

Therefore, we obtain the adversarial disentanglement loss function as follow:
\[ \mathcal L_{adv}= \mathcal L_{cross}+ \mathcal L'_{cross} \tag{14} \]

The overall training loss, namely $\mathcal{L}_{\rm EC}$ calculated as:
\[ \mathcal{L}_{\rm EC} =   \mathcal{L}_{\rm cross} + \alpha\mathcal{L}_{\rm KL}  + \beta\mathcal L_{adv} \tag{15}\]
where $\alpha and \beta$ are hyperparameter mentioned in Implementation Details.

As a result, the combined loss facilitates the model's learning of emotional continuity coding and emotional attribution coding, ultimately improving its overall performance.

\section{Experiments}
\subsection{Dataset}
We assess the performance of HCAN on three benchmark datasets which are IEMOCAP\cite{busso2008iemocap}, MELD\cite{poria2018meld} and EmoryNLP\cite{zahiri2017emotion}.

\textbf{IEMOCAP} is a dataset recorded as dyadic conversational video clips with eight speaker participating in the training set while two speaker in testing set.


\textbf{MELD} dataset is a multimodal dataset that has been expanded from the EmotionLines dataset with seven emotional labels. MELD is obtained from the popular TV show \textit{Friends} and comprises over 1400 dialogues and 13000 utterances.

\textbf{EmoryNLP} is a textual dataset also collected from the TV series \textit{Friends}. The dataset comprises utterances that are categorized into seven distinct emotional classes.

In this work, we only consider the emotional classes for the MELD and EmoryNLP datasets. Additionally, we maintain consistency with COSMIC in terms of the train/val/test splits. The details of datasets are presented in TABLE I and TABLE II.
\begin{table}[htbp]
\begin{center}
\renewcommand{\arraystretch}{1.5}
\caption{The Statistics of splits used in different datasets}
\begin{tabular}{|c|ccc|ccc|}
\hline
{\textbf{Dataset}} & \multicolumn{3}{c|}{\#Dialogue} & \multicolumn{3}{c|}{\#Utterance}                              \\ \cline{2-7} 
                                  & \multicolumn{1}{c|}{Train} & \multicolumn{1}{c|}{Val} & Test & \multicolumn{1}{c|}{Train} & \multicolumn{1}{c|}{Val}  & Test \\ \hline
\textbf{IEMOCAP}                  & \multicolumn{2}{c|}{120}                              & 31   & \multicolumn{2}{c|}{5810}                              & 1623 \\ \hline
\textbf{MELD}                     & \multicolumn{1}{c|}{1039}  & \multicolumn{1}{c|}{114} & 280  & \multicolumn{1}{c|}{9989}  & \multicolumn{1}{c|}{1109} & 2610 \\ \hline
\textbf{EmoryNLP}                 & \multicolumn{1}{c|}{659}   & \multicolumn{1}{c|}{89}  & 79   & \multicolumn{1}{c|}{7551}  & \multicolumn{1}{c|}{954}  & 984  \\ \hline
\end{tabular}
\end{center}

\begin{center}
\renewcommand{\arraystretch}{1.5}
\caption{The Statistics of evaluation metrics used in different datasets}
\begin{tabular}{|c|c|c|c|}
\hline
\textbf{Dataset}  & \# classes & Metric           & \# Speakers\\ \hline
\textbf{IEMOCAP}  & 6          & Weighted Avg. F1 & 2\\ \hline
\textbf{EmoryNLP}     & 7          & Weighted Avg. F1 & 2-3\\ \hline
\textbf{MELD} & 7          & Weighted Avg. F1         & 2-3\\ \hline
\end{tabular}
    
\end{center}
\end{table}



\subsection{Baselines}
For the baselines, we mainly select two groups of outstanding models to compare with our approach. 
\subsubsection{Recurrent-Based Methods}
\textbf{DialogueRNN}\cite{majumder2019dialoguernn} dynamically models emotions by taking into account the current speaker, contextual content, and emotional state, with a focus on distinguishing between different speakers.
\textbf{COSMIC}\cite{ghosal2020cosmic} is a conversational model that incorporates commonsense knowledge to improve its performance which injects commonsense knowledge into Gated Recurrent Units to capture the internal state, external state, and intent state' features.
\textbf{SKAIG}\cite{li2021past} is improved by incorporating action information inferred from the preceding context and the intention suggested by the subsequent context. Additionally, it utilized a CSK method to represent the edges with knowledge, and introduced a graphics converter to handle them. 
\textbf{DialogueCRN}\cite{DBLP:conf/acl/HuWH20} designs multi-turn reasoning modules to extract and integrate emotional clues which performs an iterative process of intuitive retrieval and conscious reasoning, mimicking the unique cognitive thinking of humans.

\subsubsection{Attention-Based Methods} 
\textbf{KET}\cite{zhong2019knowledge} utilizes external commonsense knowledge through the use of hierarchical self-attention and context-aware graph attention. This approach allows for dynamic incorporation of knowledge into transformers, resulting in a knowledge-enriched model.
\textbf{DAG-ERC}\cite{shen2021directed} regards the internal structure of dialogue as a directed acyclic graph to encode utterances, providing a more intuitive approach to model the information flow between the distant conversation background and the nearby context.
\textbf{TODKAT}\cite{zhu2021topic} proposes a language model (LM) augmented with topics, which includes an additional layer specialized in detecting topics, and incorporates commonsense statements obtained from a knowledge base based on the dialogue context.
\textbf{CoG-BART}\cite{li2022contrast} presents a new method that employs a contrastive loss and a task for generating responses to ensure that distinct emotions are mutually exclusive.
\subsection{Implementation Detail}

Following COSMIC\cite{ghosal2020cosmic}, we only utilize utterance-level text features that are fine-tuned using RoBERTa\cite{liu2019roberta} to accomplish the ERC task. We conduct all HCAN experiments with a learning rate of 1e-4.
The batch size is set to 32 and the dropout rate is kept at 0.2. 
The number of LSTM layers was set to 2, 1, and 1 on IEMOCAP, MELD, and EmoryNLP datasets, respectively.
The number of heads in standard multi-head attention and IA-attention are 8 and 4, respectively.
The hyperparameter $\alpha$ is set as 0.1, 0.2, 0.2 for IEMOCAP, MELD, and EmoryNLP datasets while $\beta$ is unified as 0.05.
The results reported in our experiments are based on the average score of 5 random runs on the test set. 
A server with one NVIDIA A100(40G) GPU is used to conduct our experiments.
The addtional reproduction experiments are aligned to the baselines strictly.
\begin{table}[htbp]
\begin{center}
\renewcommand{\arraystretch}{1.5}
\caption{F1 scores on three benchmark. The best results are in bold}
\begin{tabular}{|c|c|c|c|}
\hline
\textbf{Models} & \textbf{IEMOCAP} & \textbf{MELD} & \textbf{EmoryNLP} \\ \cline{2-4} 
                                 & W-Avg F1         & W-Avg F1      & W-Avg F1          \\ \hline
KET                             & 61.33            & 58.18             & 34.39                 \\ \hline
DialogueRNN$^\dagger$                       & 62.75            & -             & -                 \\ \hline
TODKAT$^\dagger$                           & 63.75            & 65.27         & 38.59             \\ \hline
DialogueGCN                     & 64.37            & 58.10             & -                 \\ \hline
COSMIC$^\dagger$                           & 65.28            & 64.21         & 37.61             \\ \hline
DialogXL                       & 66.2            & 62.41             & 34.73                 \\ \hline
DialogueCRN                    & 66.33            & 58.39            & -                 \\ \hline
SKAIG$^\dagger$                       & 65.79            & 65.18             & 37.57                \\ \hline
DAG-ERC$^\dagger$                       & 68.03            & 63.65             & 38.94                \\ \hline
COG-BART                         & 66.18            & 64.81         & 39.04             \\ \hline
DialogueRNN$^\dagger_{+ \rm EAE}$                      & 64.85{$(\uparrow{2.10})$}            & -             & -                 \\ \hline
COSMIC$^\dagger_{+ \rm EAE}$                           & 67.77{$(\uparrow{2.50})$}           & 65.73$(\uparrow{1.52})$      & 38.71$(\uparrow{1.10})$           \\ \hline
TODKAT$^\dagger_{+ \rm EAE}$                           & 64.98$(\uparrow{1.23})$            & 65.87$(\uparrow{0.60})$         & 38.92$(\uparrow{0.33})$            \\ \hline
SKAIG$^\dagger_{+ \rm EAE}$                       & 68.09$(\uparrow{2.30})$            & 65.68$(\uparrow{0.50})$            & 38.50$(\uparrow{1.07})$                 \\ \hline
DAG-ERC$^\dagger_{+ \rm EAE}$                       & 68.80$(\uparrow{0.77})$            & 64.73$(\uparrow{1.08})$            & 39.45$(\uparrow{0.51})$                 \\ \hline

HCAN(Ours)                        & \bf{69.21}            & \bf{66.24}         & \bf{39.67}             \\ \hline
\multicolumn{4}{l}{$\dagger$ indicates our reproduction results with the same settings in baselines.} \\
\multicolumn{4}{l}{$_{+ \rm EAE}$ means the model added with EAE module.} \\

\end{tabular}
\end{center}
\end{table}
\subsection{Main Result}
TABLE III shows the main results of HCAN on three benchmarks compared to previous methods.
The results demonstrate that our HCAN achieves the best performance across all three datasets. Furthermore, compared to the previous state-of-the-art (SOTA) models on IEMOCAP, MELD, and EmoryNLP, HCAN outperforms them by 1.18\%, 0.95\%, and 0.63\%, respectively. 
IEMOCAP is known for having longer multi-turn dialogues and a well-balanced distribution of emotions, which allows for a more comprehensive evaluation of model performance. Our significant improvement(1.18\%) in performance on this dataset successfully demonstrates the model's ability to model long-distance emotional continuity and effectiveness in dyadic conversational scenario.
MELD and EmoryNLP datasets consist of multiple dialogue roles and shorter conversations, which closely resemble real-life scenarios. Additionally, these datasets have highly imbalanced emotion categories. Our model's improvement on these datasets demonstrates its effectiveness in capturing complex dialogue relationships and interpersonal emotional dependencies, as well as its robustness in recognizing different emotions.
It is worth noting that the previous SOTA models were achieved using different models for each dataset, as the sample characteristics of each dataset vary significantly. However, our method unifies the SOTA across these benchmarks, demonstrating the generalizability of our approach in different application scenarios.

\subsection{Ablation Studies}
As shown in TBALE IV, we conducted more detailed ablation experiments to quantify the contributions of the ECE module, EAE module,  ${\mathcal{L}_{\rm KL}}$, ${\mathcal{L}_{\rm sec}}$ to the performance.
(1) For ECE module, the ablation experiments leads to a performance decrease of 2.75\%, 0.60\%and  0.32\% on IEMOCAP, MELD and EmoryNLP respectively, demonstrating its generalization on different scenarios and especially effectiveness in long conversation.
(2) For EAE module, the removal of EAE leads to a performance decrease of 0.67\%, 1.65\% and 1.99\%on IEMOCAP, MELD and EmoryNLP respectively. The results elaborate the effectiveness of EAE and the importance of emotional attribution modeling based on dialogue relationship.
(3) For KL loss, the removal of $\mathcal{L}_{\rm KL}$ causes a decrease in model performance by 0.93\% on the EmoryNLP dataset. This suggests the effectiveness of KL in detecting emotional shifts, as this dataset often contains emotional shifting samples.
Overall, the unique contributions of different modules jointly contribute to the generalization and effectiveness of HCAN.


\begin{table}[htbp]
\begin{center}
\renewcommand{\arraystretch}{1.5}
\caption{Ablation experiments of HCAN's components on three different benchmarks}
\begin{tabular}{|c|c|c|c|}
\hline
\textbf{Models} & \textbf{IEMOCAP} & \textbf{MELD} & \textbf{EmoryNLP} \\ \cline{2-4} 
                                 & W-Avg F1         & W-Avg F1      & W-Avg F1          \\ \hline
HCAN                        & \bf{69.21}            & \bf{66.24}         & \bf{39.67}             \\ \hline
- w/o ECE                           & 66.46$(\downarrow{2.75})$            & 65.73$(\downarrow{0.60})$         & 38.95$(\downarrow{0.72})$             \\ \hline
- w/o EAE                            & 68.57{$(\downarrow{0.67})$}           & 64.59$(\downarrow{1.65})$         & 37.28$(\downarrow{2.39})$            \\ \hline
- w/o ${\mathcal{L}_{\rm KL}}$     & 69.13$(\downarrow{0.08})$           & 65.97$(\downarrow{0.27})$         & 38.74$(\downarrow{0.93})$             \\ \hline
\multicolumn{4}{l}{- w/o * indicates the experimental results without the * module in HCAN} \\

\end{tabular}
\end{center}
\end{table}

\subsection{The Exploration of Generality}
 Regarding the universality of the EAE module, as it has strong transferability, we conducted experiments by adding it to different models based on recurrent and attention-based methods shown in TBALE II. The results show that our EAE module can effectively improve the performance of models based on different architectures. Moreover, the performance improvement on IEMOCAP, a dataset with long dialogues, is stronger than that on MELD, which has shorter conversations. Meanwhile, we observe that the improvement in models based on recurrent methods(i.e. COSMIC) is greater than that in models based on attention mechanisms(i.e TODKAT). This is logical because our EAE module is implemented based on attention mechanisms, which are naturally superior to temporal structures in modeling various levels of emotional attribution. It is reasonable to assume that attention-based methods implicitly capture emotional attribution to some extent, while our method captures more comprehensive emotional attribution information, leading to performance improvement.

\begin{table}[htbp]
\begin{center}
\renewcommand{\arraystretch}{1.5}
\caption{Ablation Experiment of $\mathcal{L}_{adv}$ : F1 scores on three benchmark}
\begin{tabular}{|c|c|c|c|}
\hline
\textbf{Models} & \textbf{IEMOCAP} & \textbf{MELD} & \textbf{EmoryNLP} \\ \cline{2-4} 
                                 & W-Avg F1         & W-Avg F1      & W-Avg F1          \\ \hline
DialogueRNN$^\dagger$                       & 62.75            & -             & -                 \\ \hline
TODKAT$^\dagger$                           & 63.75            & 65.27         & 38.59             \\ \hline
COSMIC$^\dagger$                           & 65.28            & 64.21         & 37.61             \\ \hline
SKAIG$^\dagger$                       & 65.79            & 65.18             & 37.57                \\ \hline
DAG-ERC$^\dagger$                       & 68.03            & 63.65             & 38.94                \\ \hline
DialogueRNN$^\dagger_{+ \mathcal{L}_{adv}}$                      & 63.61            & -             & -                 \\ \hline
COSMIC$^\dagger_{+ \mathcal{L}_{adv}}$                           & 66.54           & 65.28      & 38.60           \\ \hline
TODKAT$^\dagger_{+ \mathcal{L}_{adv}}$                           & 64.12           & 65.54         & 38.78            \\ \hline
SKAIG$^\dagger_{+ \mathcal{L}_{adv}}$                       & 67.28           & 65.46            & 38.39                 \\ \hline
DAG-ERC$^\dagger_{+ \mathcal{L}_{adv}}$                       & 68.40            & 64.73            & 39.12                 \\ \hline

HCAN(Ours)$_{- \mathcal{L}_{adv}}$                         & {69.03}            & {65.92}         & {39.29}             \\ \hline
HCAN(Ours)                        & \bf{69.21}            & \bf{66.24}         & \bf{39.67}             \\ \hline
\multicolumn{4}{l}{$\dagger$ indicates our reproduction results with the same settings in baselines.} \\
\multicolumn{4}{l}{$_{+ \mathcal{L}_{adv}}$ means the model added with $_{+ \mathcal{L}_{adv}}$.} \\

\end{tabular}
\end{center}
\end{table}

 \subsection{The Robustness of Speaker Modeling}
By incorporating the ECE module to capture the conversational dynamics, our model has successfully captured rich speaker characteristics. 
Our approach to modeling speaker robustness is primarily reflected in the \textbf{Adversarial Emotion Disentanglement} loss.
To quantify the contribution of this loss in mitigating speaker modeling overfitting, we conducted experiments similar to those in the EAE module's generalization study.

TBALE V shows that removing the ${ \mathcal{L}_{adv}}$ module results in a certain degree of performance degradation for the HCAN model. Conversely, adding the $_{+ \mathcal{L}_{adv}}$ module to other baselines leads to significant performance improvements. For the SKAIG and COSMIC models, which utilize a large amount of common sense knowledge to model speaker emotions, our loss function effectively prevents overfitting on the IEMOCAP and MELD datasets, while maintaining their performance improvements. However, for models that focus on modeling conversational dynamics, such as DAG-ERC, the effect of loss improvement is limited. This is because their modeling of conversational dynamics enhances the robustness of speaker modeling to some extent.


\begin{figure}[htbp]
\centerline{\includegraphics[scale=0.75]{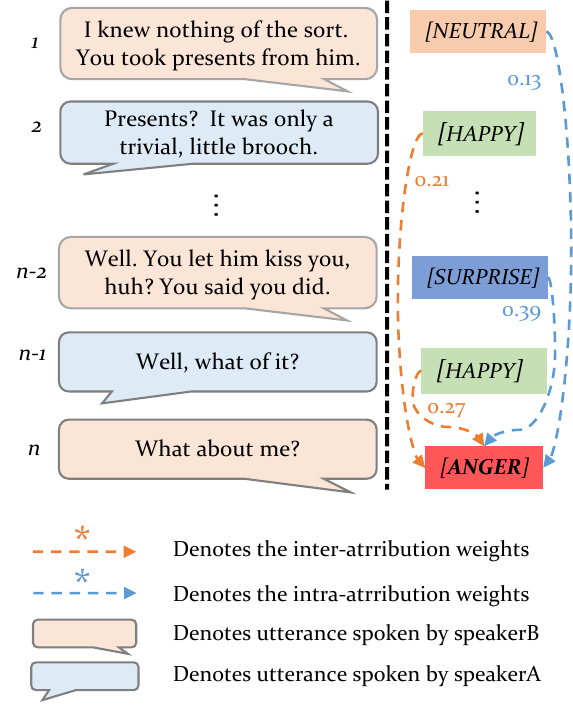}}
\caption{Dyadic conversation for case study.}
\label{fig:case}
\end{figure}

\subsection{Case Study}
Fig.~\ref{fig:case} shows a segment of a dyadic conversation. Intuitively, the anger expressed by speakerB in the $n^{th}$ sentence seems to have been mainly triggered by his own surprise towards ``kiss'' and the question posed by speakerA in the ${n\text{-}1}^{th}$ sentence. Meanwhile, the perfunctory response from speakerA in the $2^{rd}$ sentence may have also contributed to some extent.It is evident that the distribution of predictive emotion aligns with that of identifying the emotion, and the attention weights further demonstrate model's ability to effectively capture the relationship in long-distance conversations.

\section{Conclusion}
Insufficient modeling of context and ambiguous capture of dialogue relationships have been persistent challenges in improving the performance of ERC models. In this work, we propose HCAN to significantly addresses these issues.
Our proposed ECE module achieves strong robustness in modeling global emotion continuity across different datasets by combining recurrent and attention-based approaches. 
It particularly demonstrates outstanding performance on long conversation samples. 
Meanwhile, the proposed EAE module extracts intra-attribution and inter-attribution features, which offers a more direct and accurate modeling of human emotional comprehension in the perspective of Attribution Theory.
The proposed comprehensive loss function, namely Emotional Cognitive Loss $\mathcal{L}_{\rm EC}$, which effectively mitigates emotional drift and addresses the issue of overfitting in speaker modeling.
Moreover, EAE module exhibits strong generalization and effectiveness when added to current models.
Our model achieves state-of-the-art performance on three datasets, demonstrating the superiority of our work.

\bibliographystyle{IEEEtran}
\bibliography{ICTAI.bib}

\begin{thebibliography}{10}
\providecommand{\url}[1]{#1}
\csname url@samestyle\endcsname
\providecommand{\newblock}{\relax}
\providecommand{\bibinfo}[2]{#2}
\providecommand{\BIBentrySTDinterwordspacing}{\spaceskip=0pt\relax}
\providecommand{\BIBentryALTinterwordstretchfactor}{4}
\providecommand{\BIBentryALTinterwordspacing}{\spaceskip=\fontdimen2\font plus
\BIBentryALTinterwordstretchfactor\fontdimen3\font minus
  \fontdimen4\font\relax}
\providecommand{\BIBforeignlanguage}[2]{{%
\expandafter\ifx\csname l@#1\endcsname\relax
\typeout{** WARNING: IEEEtran.bst: No hyphenation pattern has been}%
\typeout{** loaded for the language `#1'. Using the pattern for}%
\typeout{** the default language instead.}%
\else
\language=\csname l@#1\endcsname
\fi
#2}}
\providecommand{\BIBdecl}{\relax}
\BIBdecl

\bibitem{munikar2019fine}
M.~Munikar, S.~Shakya, and A.~Shrestha, ``Fine-grained sentiment classification
  using bert,'' in \emph{2019 Artificial Intelligence for Transforming Business
  and Society (AITB)}, vol.~1.\hskip 1em plus 0.5em minus 0.4em\relax IEEE,
  2019, pp. 1--5.

\bibitem{yin2020sentibert}
D.~Yin, T.~Meng, and K.-W. Chang, ``Sentibert: A transferable transformer-based
  architecture for compositional sentiment semantics,'' \emph{arXiv preprint
  arXiv:2005.04114}, 2020.

\bibitem{do2019deep}
H.~H. Do, P.~W. Prasad, A.~Maag, and A.~Alsadoon, ``Deep learning for
  aspect-based sentiment analysis: a comparative review,'' \emph{Expert systems
  with applications}, vol. 118, pp. 272--299, 2019.

\bibitem{sun2019utilizing}
C.~Sun, L.~Huang, and X.~Qiu, ``Utilizing bert for aspect-based sentiment
  analysis via constructing auxiliary sentence,'' in \emph{Proceedings of
  NAACL-HLT}, 2019, pp. 380--385.

\bibitem{cambria2017sentiment}
E.~Cambria, S.~Poria, A.~Gelbukh, and M.~Thelwall, ``Sentiment analysis is a
  big suitcase,'' \emph{IEEE Intelligent Systems}, vol.~32, no.~6, pp. 74--80,
  2017.

\bibitem{anstead2015social}
N.~Anstead and B.~O'Loughlin, ``Social media analysis and public opinion: The
  2010 uk general election,'' \emph{Journal of computer-mediated
  communication}, vol.~20, no.~2, pp. 204--220, 2015.

\bibitem{sheridan2016human}
T.~B. Sheridan, ``Human--robot interaction: status and challenges,''
  \emph{Human factors}, vol.~58, no.~4, pp. 525--532, 2016.

\bibitem{rashkin2018towards}
H.~Rashkin, E.~M. Smith, M.~Li, and Y.-L. Boureau, ``Towards empathetic
  open-domain conversation models: A new benchmark and dataset,'' \emph{arXiv
  preprint arXiv:1811.00207}, 2018.

\bibitem{qixiang-etal-2022-exploiting}
\BIBentryALTinterwordspacing
G.~Qixiang, G.~Dong, Y.~Mou, L.~Wang, C.~Zeng, D.~Guo, M.~Sun, and W.~Xu,
  ``Exploiting domain-slot related keywords description for few-shot
  cross-domain dialogue state tracking,'' in \emph{Proceedings of the 2022
  Conference on Empirical Methods in Natural Language Processing}.\hskip 1em
  plus 0.5em minus 0.4em\relax Abu Dhabi, United Arab Emirates: Association for
  Computational Linguistics, Dec. 2022, pp. 2460--2465. [Online]. Available:
  \url{https://aclanthology.org/2022.emnlp-main.157}
\BIBentrySTDinterwordspacing

\bibitem{zeng2022semi}
W.~Zeng, K.~He, Z.~Wang, D.~Fu, G.~Dong, R.~Geng, P.~Wang, J.~Wang, C.~Sun,
  W.~Wu \emph{et~al.}, ``Semi-supervised knowledge-grounded pre-training for
  task-oriented dialog systems,'' \emph{arXiv preprint arXiv:2210.08873}, 2022.

\bibitem{sun2023improving}
M.~Sun, Q.~Gao, Y.~Mou, G.~Dong, R.~Liu, and W.~Guo, ``Improving few-shot
  performance of dst model through multitask to better serve language-impaired
  people,'' in \emph{2023 IEEE International Conference on Acoustics, Speech,
  and Signal Processing Workshops (ICASSPW)}.\hskip 1em plus 0.5em minus
  0.4em\relax IEEE, 2023, pp. 1--5.

\bibitem{shen2021directed}
W.~Shen, S.~Wu, Y.~Yang, and X.~Quan, ``Directed acyclic graph network for
  conversational emotion recognition,'' \emph{arXiv preprint arXiv:2105.12907},
  2021.

\bibitem{majumder2019dialoguernn}
N.~Majumder, S.~Poria, D.~Hazarika, R.~Mihalcea, A.~Gelbukh, and E.~Cambria,
  ``Dialoguernn: An attentive rnn for emotion detection in conversations,'' in
  \emph{Proceedings of the AAAI conference on artificial intelligence},
  vol.~33, no.~01, 2019, pp. 6818--6825.

\bibitem{ghosal2020cosmic}
D.~Ghosal, N.~Majumder, A.~Gelbukh, R.~Mihalcea, and S.~Poria, ``Cosmic:
  Commonsense knowledge for emotion identification in conversations,''
  \emph{arXiv preprint arXiv:2010.02795}, 2020.

\bibitem{hazarika2018icon}
D.~Hazarika, S.~Poria, R.~Mihalcea, E.~Cambria, and R.~Zimmermann, ``Icon:
  Interactive conversational memory network for multimodal emotion detection,''
  in \emph{Proceedings of the 2018 conference on empirical methods in natural
  language processing}, 2018, pp. 2594--2604.

\bibitem{jiao2019higru}
W.~Jiao, H.~Yang, I.~King, and M.~R. Lyu, ``Higru: Hierarchical gated recurrent
  units for utterance-level emotion recognition,'' \emph{arXiv preprint
  arXiv:1904.04446}, 2019.

\bibitem{zhong2019knowledge}
P.~Zhong, D.~Wang, and C.~Miao, ``Knowledge-enriched transformer for emotion
  detection in textual conversations,'' \emph{arXiv preprint arXiv:1909.10681},
  2019.

\bibitem{ghosal2019dialoguegcn}
D.~Ghosal, N.~Majumder, S.~Poria, N.~Chhaya, and A.~Gelbukh, ``Dialoguegcn: A
  graph convolutional neural network for emotion recognition in conversation,''
  \emph{arXiv preprint arXiv:1908.11540}, 2019.

\bibitem{shen2021dialogxl}
W.~Shen, J.~Chen, X.~Quan, and Z.~Xie, ``Dialogxl: All-in-one xlnet for
  multi-party conversation emotion recognition,'' in \emph{Proceedings of the
  AAAI Conference on Artificial Intelligence}, vol.~35, no.~15, 2021, pp.
  13\,789--13\,797.

\bibitem{li2022contrast}
S.~Li, H.~Yan, and X.~Qiu, ``Contrast and generation make bart a good dialogue
  emotion recognizer,'' in \emph{Proceedings of the AAAI conference on
  artificial intelligence}, vol.~36, no.~10, 2022, pp. 11\,002--11\,010.

\bibitem{chen2023dialogue}
H.~Chen, P.~Hong, W.~Han, N.~Majumder, and S.~Poria, ``Dialogue relation
  extraction with document-level heterogeneous graph attention networks,''
  \emph{Cognitive Computation}, pp. 1--10, 2023.

\bibitem{sun2021discourse}
Y.~Sun, N.~Yu, and G.~Fu, ``A discourse-aware graph neural network for emotion
  recognition in multi-party conversation,'' in \emph{Findings of the
  Association for Computational Linguistics: EMNLP 2021}, 2021, pp. 2949--2958.

\bibitem{dong2023multi}
G.~Dong, Z.~Wang, J.~Zhao, G.~Zhao, D.~Guo, D.~Fu, T.~Hui, C.~Zeng, K.~He,
  X.~Li \emph{et~al.}, ``A multi-task semantic decomposition framework with
  task-specific pre-training for few-shot ner,'' \emph{arXiv preprint
  arXiv:2308.14533}, 2023.

\bibitem{liu2020coach}
Z.~Liu, G.~I. Winata, P.~Xu, and P.~Fung, ``Coach: A coarse-to-fine approach
  for cross-domain slot filling,'' \emph{arXiv preprint arXiv:2004.11727},
  2020.

\bibitem{schachter1962cognitive}
S.~Schachter and J.~Singer, ``Cognitive, social, and physiological determinants
  of emotional state.'' \emph{Psychological review}, vol.~69, no.~5, p. 379,
  1962.

\bibitem{dong2022pssat}
G.~Dong, D.~Guo, L.~Wang, X.~Li, Z.~Wang, C.~Zeng, K.~He, J.~Zhao, H.~Lei,
  X.~Cui \emph{et~al.}, ``Pssat: A perturbed semantic structure awareness
  transferring method for perturbation-robust slot filling,'' \emph{arXiv
  preprint arXiv:2208.11508}, 2022.

\bibitem{guo2023revisit}
D.~Guo, G.~Dong, D.~Fu, Y.~Wu, C.~Zeng, T.~Hui, L.~Wang, X.~Li, Z.~Wang, K.~He
  \emph{et~al.}, ``Revisit out-of-vocabulary problem for slot filling: A
  unified contrastive framework with multi-level data augmentations,'' in
  \emph{ICASSP 2023-2023 IEEE International Conference on Acoustics, Speech and
  Signal Processing (ICASSP)}.\hskip 1em plus 0.5em minus 0.4em\relax IEEE,
  2023, pp. 1--5.

\bibitem{li2022robust}
X.~Li, H.~Lei, L.~Wang, G.~Dong, J.~Zhao, J.~Liu, W.~Xu, and C.~Zhang, ``A
  robust contrastive alignment method for multi-domain text classification,''
  in \emph{ICASSP 2022-2022 IEEE International Conference on Acoustics, Speech
  and Signal Processing (ICASSP)}.\hskip 1em plus 0.5em minus 0.4em\relax IEEE,
  2022, pp. 7827--7831.

\bibitem{DBLP:conf/acl/HuWH20}
D.~Hu, L.~Wei, and X.~Huai, ``Dialoguecrn: Contextual reasoning networks for
  emotion recognition in conversations,'' in \emph{{ACL/IJCNLP} {(1)}}.\hskip
  1em plus 0.5em minus 0.4em\relax Association for Computational Linguistics,
  2021, pp. 7042--7052.

\bibitem{zhao2022cauain}
W.~Zhao, Y.~Zhao, and X.~Lu, ``Cauain: Causal aware interaction network for
  emotion recognition in conversations,'' in \emph{Proceedings of the
  Thirty-First International Joint Conference on Artificial Intelligence,
  IJCAI}, 2022, pp. 4524--4530.

\bibitem{zhu2021topic}
L.~Zhu, G.~Pergola, L.~Gui, D.~Zhou, and Y.~He, ``Topic-driven and
  knowledge-aware transformer for dialogue emotion detection,'' \emph{arXiv
  preprint arXiv:2106.01071}, 2021.

\bibitem{yu2020dialogue}
D.~Yu, K.~Sun, C.~Cardie, and D.~Yu, ``Dialogue-based relation extraction,''
  \emph{arXiv preprint arXiv:2004.08056}, 2020.

\bibitem{vaswani2017attention}
A.~Vaswani, N.~Shazeer, N.~Parmar, J.~Uszkoreit, L.~Jones, A.~N. Gomez,
  {\L}.~Kaiser, and I.~Polosukhin, ``Attention is all you need,''
  \emph{Advances in neural information processing systems}, vol.~30, 2017.

\bibitem{dong2023bridging}
G.~Dong, R.~Li, S.~Wang, Y.~Zhang, Y.~Xian, and W.~Xu, ``Bridging the kb-text
  gap: Leveraging structured knowledge-aware pre-training for kbqa,''
  \emph{arXiv preprint arXiv:2308.14436}, 2023.

\bibitem{zhao2022entity}
G.~Zhao, G.~Dong, Y.~Shi, H.~Yan, W.~Xu, and S.~Li, ``Entity-level interaction
  via heterogeneous graph for multimodal named entity recognition,'' in
  \emph{Findings of the Association for Computational Linguistics: EMNLP 2022},
  2022, pp. 6345--6350.

\bibitem{guo2019gaussian}
M.~Guo, Y.~Zhang, and T.~Liu, ``Gaussian transformer: a lightweight approach
  for natural language inference,'' in \emph{Proceedings of the AAAI Conference
  on Artificial Intelligence}, vol.~33, no.~01, 2019, pp. 6489--6496.

\bibitem{li2023generative}
X.~Li, L.~Wang, G.~Dong, K.~He, J.~Zhao, H.~Lei, J.~Liu, and W.~Xu,
  ``Generative zero-shot prompt learning for cross-domain slot filling with
  inverse prompting,'' \emph{arXiv preprint arXiv:2307.02830}, 2023.

\bibitem{dong2023prototypical}
G.~Dong, Z.~Wang, L.~Wang, D.~Guo, D.~Fu, Y.~Wu, C.~Zeng, X.~Li, T.~Hui, K.~He
  \emph{et~al.}, ``A prototypical semantic decoupling method via joint
  contrastive learning for few-shot named entity recognition,'' in \emph{ICASSP
  2023-2023 IEEE International Conference on Acoustics, Speech and Signal
  Processing (ICASSP)}.\hskip 1em plus 0.5em minus 0.4em\relax IEEE, 2023, pp.
  1--5.

\bibitem{goodfellow2014explaining}
I.~J. Goodfellow, J.~Shlens, and C.~Szegedy, ``Explaining and harnessing
  adversarial examples,'' \emph{arXiv preprint arXiv:1412.6572}, 2014.

\bibitem{kurakin2016adversarial}
A.~Kurakin, I.~J. Goodfellow, and S.~Bengio, ``Adversarial examples in the
  physical world,'' \emph{arXiv preprint arXiv:1607.02533}, 2016.

\bibitem{manzini2019black}
T.~Manzini, Y.~C. Lim, Y.~Tsvetkov, and A.~W. Black, ``Black is to criminal as
  caucasian is to police: Detecting and removing multiclass bias in word
  embeddings,'' \emph{arXiv preprint arXiv:1904.04047}, 2019.

\bibitem{busso2008iemocap}
C.~Busso, M.~Bulut, C.-C. Lee, A.~Kazemzadeh, E.~Mower, S.~Kim, J.~N. Chang,
  S.~Lee, and S.~S. Narayanan, ``Iemocap: Interactive emotional dyadic motion
  capture database,'' \emph{Language resources and evaluation}, vol.~42, pp.
  335--359, 2008.

\bibitem{poria2018meld}
S.~Poria, D.~Hazarika, N.~Majumder, G.~Naik, E.~Cambria, and R.~Mihalcea,
  ``Meld: A multimodal multi-party dataset for emotion recognition in
  conversations,'' \emph{arXiv preprint arXiv:1810.02508}, 2018.

\bibitem{zahiri2017emotion}
S.~M. Zahiri and J.~D. Choi, ``Emotion detection on tv show transcripts with
  sequence-based convolutional neural networks,'' \emph{arXiv preprint
  arXiv:1708.04299}, 2017.

\bibitem{li2021past}
J.~Li, Z.~Lin, P.~Fu, and W.~Wang, ``Past, present, and future: Conversational
  emotion recognition through structural modeling of psychological knowledge,''
  in \emph{Findings of the association for computational linguistics: EMNLP
  2021}, 2021, pp. 1204--1214.

\bibitem{liu2019roberta}
Y.~Liu, M.~Ott, N.~Goyal, J.~Du, M.~Joshi, D.~Chen, O.~Levy, M.~Lewis,
  L.~Zettlemoyer, and V.~Stoyanov, ``Roberta: A robustly optimized bert
  pretraining approach,'' \emph{arXiv preprint arXiv:1907.11692}, 2019.

\end{thebibliography}

\end{document}